\documentclass[letterpaper, 10 pt, conference]{ieeeconf}  % Comment this line out if you need a4paper

\IEEEoverridecommandlockouts                              % This command is only needed if 
\usepackage{graphicx} % for pdf, bitmapped graphics files
\usepackage{epsfig} % for postscript graphics files
\usepackage{mathptmx} % assumes new font selection scheme installed
\usepackage{times} % assumes new font selection scheme installed
\usepackage{amsmath} % assumes amsmath package installed
\usepackage{amssymb}  % assumes amsmath package installed
\usepackage{multirow}

\usepackage{float}

\floatstyle{ruled}
\newfloat{algorithm}{tbp}{loa}
\floatname{algorithm}{Algorithm}
\newcounter{algline}
\newcounter{algindent}
\newenvironment{algorithmic}[1][1]{%
  \setcounter{algline}{0}%
  \setcounter{algindent}{0}%
  \begin{list}{}{\setlength{\leftmargin}{3.2em}%
                 \setlength{\labelwidth}{2.7em}%
                 \setlength{\labelsep}{0.5em}%
                 \setlength{\itemsep}{0pt}%
                 \setlength{\parsep}{0pt}%
                 \setlength{\topsep}{2pt}}%
}{\end{list}}
\newcommand{\State}{\stepcounter{algline}\item[\scriptsize\arabic{algline}:]\hspace*{\value{algindent}em}}
\newcommand{\Require}{\item[\textbf{Input:}]}
\newcommand{\Ensure}{\item[\textbf{Output:}]}
\newcommand{\For}[1]{\State\textbf{for} #1 \textbf{do}\addtocounter{algindent}{1}}
\newcommand{\EndFor}{\addtocounter{algindent}{-1}\State\textbf{end for}}
\newcommand{\If}[1]{\State\textbf{if} #1 \textbf{then}\addtocounter{algindent}{1}}
\newcommand{\EndIf}{\addtocounter{algindent}{-1}\State\textbf{end if}}
\newcommand{\Return}{\textbf{return}\ }

\title{\LARGE \bf
Geometry vs Structure: Graph-Based Diagnostics for LiDAR Point-Cloud Simulation Fidelity}

\author{Ghazal Farhani$^{*}$ and Taufiq Rahman%
\thanks{The authors are with the Automotive and Surface Transportation
Research Centre, National Research Council Canada, London, Ontario, Canada.}%
\thanks{$^{*}$Corresponding author: 
(\texttt{ghazal.farhani@nrc-cnrc.gc.ca}).}%
}

\begin{document}

\maketitle
\thispagestyle{empty}
\pagestyle{empty}

%%%%%%%%%%%%%%%%%%%%%%%%%%%%%%%%%%%%%%%%%%%%%%%%%%%%%%%%%%%%%%%%%%%%%%%%%%%%%%%%
\begin{abstract}
Digital twins offer a scalable, cost-effective complement to real-world testing for validating autonomous driving and ADAS sensor pipelines, but quantifying their fidelity remains challenging, particularly for 3D LiDAR point clouds, where conventional geometric metrics can miss structural discrepancies. We propose a graph-based framework for evaluating the structural fidelity of simulated LiDAR point clouds against real-world scans. Scan-level metrics such as Chamfer Distance capture point-wise geometric similarity but overlook connectivity, topology, and object-level structure. We address this by constructing graphs from real and simulated point clouds, applying Louvain community detection to extract spatially coherent object-level subgraphs, and matching corresponding communities via centroid proximity. For each matched pair, we compute ($r_\lambda$), a bounded spectral metric derived from Weyl's inequality, and compare it against density-aware Chamfer Distance (CDC) as a geometric baseline. Controlled perturbation experiments show ($r_\lambda$) is invariant to rigid transformations and robust to sensor noise, while remaining sensitive to structural deformation. We validate the framework on 50 real-simulated LiDAR scan pairs (CARLA vs. Velodyne VLP-32C) spanning over 1,000 object-level communities across four representative classes: cars, vegetation, trees, and building walls. Results show geometric and structural fidelity capture complementary aspects of simulation quality, establishing graph-spectral metrics as a diagnostic layer for digital twin validation in ADAS and autonomous driving.
\end{abstract}

%%%%%%%%%%%%%%%%%%%%%%%%%%%%%%%%%%%%%%%%%%%%%%%%%%%%%%%%%%%%%%%%%%%%%%%%%%%%%%%%

\section{INTRODUCTION} \label{sec:intro}
As Advanced Driver Assistance Systems (ADAS) become increasingly 
integrated into transportation, rigorous safety testing is essential. 
Virtual testing environments (VTEs) emulate real-world scenarios in a 
controlled, cost-efficient, and scalable manner~\cite{zhang2021test}. A 
VTE must closely replicate the Operational Design Domain (ODD) 
with measurable deviations, since the ODD defines the safety boundaries 
of ADAS features~\cite{saej3016}. Despite recent advances in VTE 
development, significant gaps remain in simulation model validation and 
in establishing reliable comparison metrics, particularly for 3D LiDAR 
systems~\cite{sanz2018accuracy, ali2026comprehensive}. Most model-free 
validation methods for point clouds report global geometric metrics per 
scan, which cannot identify which specific objects fail to meet 
fidelity requirements or diagnose the nature of structural failures.

A perfect simulation would generate point clouds identical to 
real-world data; given inevitable sensor discrepancies, however, 
reliable similarity metrics are needed to quantify these differences. 
Traditional methods such as Iterative Closest Point (ICP), histogram 
matching, and Chamfer Distance measure point-wise positioning 
error but are insensitive to structural properties such as object 
topology, surface connectivity, and spatial relationships between 
components. This limitation is inherent to 3D analysis and has no 
direct counterpart in RGB camera-based methods: in point clouds, 
geometric and structural fidelity are complementary rather than 
redundant properties. Deep-learning-based segmentation could identify 
individual objects, but requires large labelled datasets and introduces 
training-data bias, making it unsuitable as a ground-truth validation 
metric. Even learned lidar retrievals equipped with principled uncertainty estimates, such as Bayesian neural networks \cite{farhani2023bayesian}, remain conditioned on their training distribution, which is precisely the dependence a ground-truth validation metric must avoid. Graph-based representations instead rely on first-principle 
structural and geometric relationships (point proximity and 
connectivity) providing interpretable, domain-agnostic structural 
analysis.

We propose a graph-based evaluation framework that complements existing 
geometric metrics by quantifying structural fidelity in simulated LiDAR 
point clouds. We represent point clouds as graphs, with nodes 
corresponding to points and edges encoding spatial proximity, and apply 
community detection to partition each graph into spatially coherent 
subgraphs representing individual objects (e.g., vehicles, trees). This 
yields object-level diagnostics that capture connectivity and 
topological relationships. For each detected community, we perform 
spectral analysis of the graph Laplacian to quantify structural 
differences via eigenvalue comparison, which admits theoretical upper 
bounds on divergence and enables principled threshold 
selection, unlike unbounded distance metrics.

\begin{figure}
    \centering
    \includegraphics[width=1\linewidth]{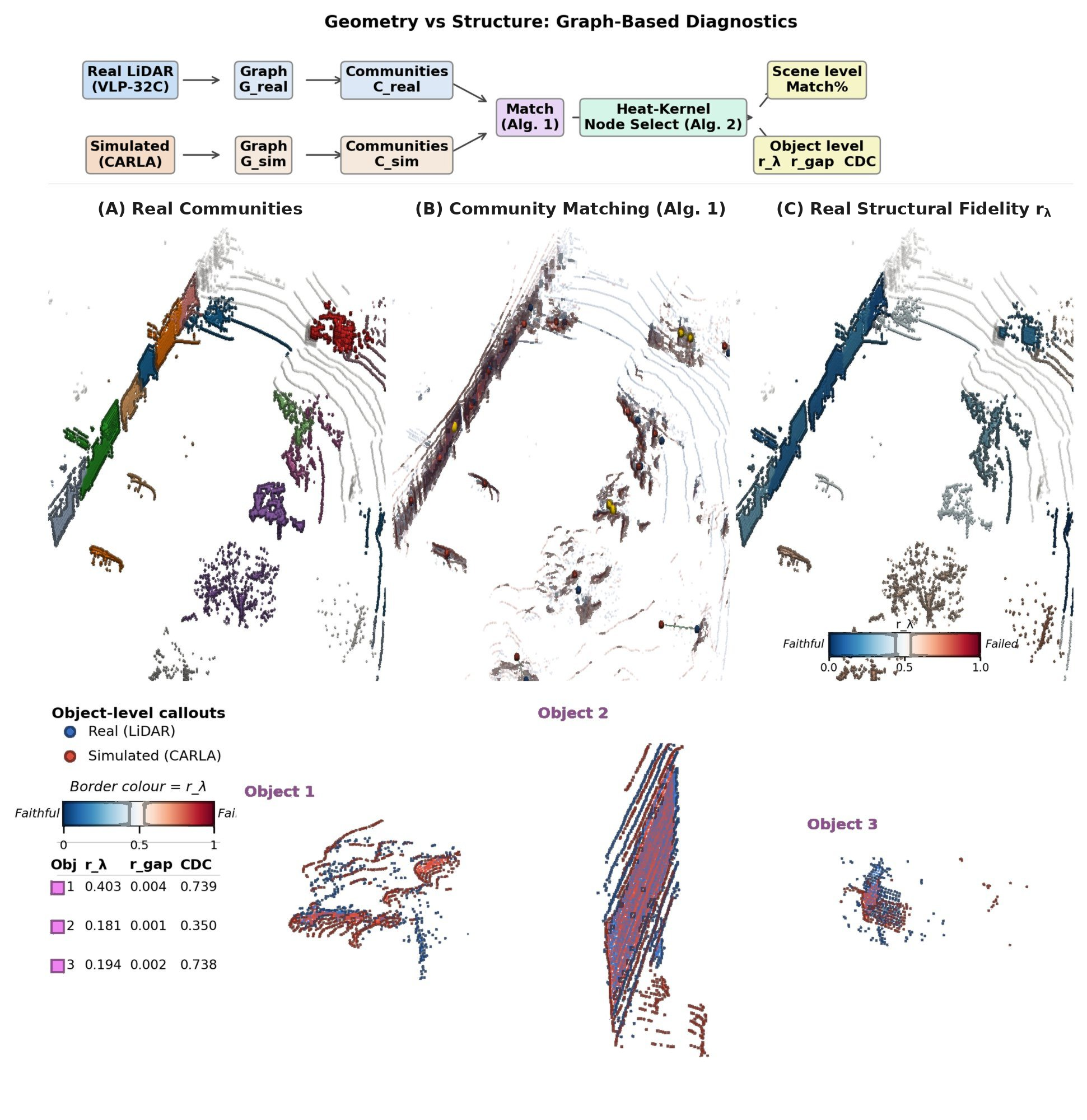}
    \caption{Overview of the proposed framework. \textit{Top:} pipeline 
    schematic from real and simulated LiDAR inputs through graph 
    construction, community detection, and matching to scene- and 
    object-level metrics. \textit{Bottom:} (A) community segmentation of 
    the real point cloud; (B) overlay of real and simulated point clouds 
    with matched community centroids; (C) structural fidelity map 
    ($r_\lambda$, blue\,=\,faithful, red\,=\,failed). The lower panels 
    show three zoomed object pairs with their geometric (CDC) and 
    structural ($r_\lambda$, $r_\mathrm{gap}$) metric values.}
    \label{fig:intro}
\end{figure}

Our framework further addresses mismatched community sizes arising from 
density differences between real and simulated sensors: we propose a 
heat kernel-based node selection method that identifies structurally 
influential points within each community, handling density mismatch 
through influence-based subgraph reduction. Although graph construction 
and eigenvalue decomposition introduce computational overhead, 
community-based partitioning enables parallel processing across 
objects. We validate the framework through controlled perturbation 
analysis and real-versus-simulated comparisons using point clouds from 
the CARLA simulator and a Velodyne VLP-32C sensor. Our results show 
that geometric and structural metrics capture complementary aspects of 
simulation fidelity, as summarized in Fig.~\ref{fig:intro}.

%%%%%%%%%%%%%%%%%%%%%%%%%%%%%%%%%%%%%%%%%%%%%%%%%%%%%%%%%%%%%%%%%%%%%%%%%%%%%%%%
\section{Related Work}\label{sec:prior_work}

Chamfer Distance and Earth Mover's Distance (EMD) are the most commonly used metrics for point cloud comparison \cite{achlioptas2018learning}. Chamfer Distance employs a nearest-neighbor approach, making it computationally efficient and well-suited for datasets with varying numbers of points. Given two sets \( A, B \subseteq \mathbb{R}^d \), the Chamfer Distance is defined as:  
\begin{equation*}
    d_{CD}(A, B) = \sum_{a \in A} \min_{b \in B} d_X(a,b),
\end{equation*}  
where \( d_X \) represents a distance measure, such as the Euclidean distance, between two points. The Chamfer Distance ranges from zero to infinity, with smaller values indicating greater similarity between the two sets.

EMD calculates the minimum cost of a one-to-one transportation problem between two point clouds:  
\begin{equation*}
    d_{\text{EMD}}(A, B) = \min_{\pi : A \to B} \sum_{a \in A} d(a, \pi(a)),
\end{equation*}  
where \( \pi \) represents a bijective mapping between \( A \) and \( B \) \cite{rubner2000earth}. Although EMD can provide more accurate geometric matching than Chamfer Distance, its \( O(n^3 \log n) \) computational complexity makes it impractical for large-scale automotive point clouds. More importantly, neither Chamfer Distance nor EMD provides theoretical bounds on structural divergence, reducing their effectiveness for detecting subtle topological variations \cite{achlioptas2018learning}.

Recently, \textit{Wu et al. 2021} \cite{wu2021density} introduced the density-aware Chamfer distance (CDC) which is formulated using an exponential-kernel transformation of the point-wise Chamfer distance. This transformation maps each point-level distance to the interval $[0,1]$, the CDC between two point sets $S_1$ and $S_2$ is approximated as
\begin{equation}
\begin{aligned}
d_{\mathrm{CDC}}(S_1,S_2)
\approx {}&
\frac{1}{|S_1|}
\sum_{x\in S_1}
\min_{y\in S_2}
\Bigl(1-e^{-\|x-y\|^2}\Bigr)
\\
&+
\frac{1}{|S_2|}
\sum_{y\in S_2}
\min_{x\in S_1}
\Bigl(1-e^{-\|y-x\|^2}\Bigr).
\end{aligned}
\label{eq:cdc}
\end{equation}
As the kernel in CDC is bounded, compared to Chamfer Distance the metric is less sensitive to extreme outlier distance.

Histogram-based methods offer an alternative approach, computing 
pairwise distances between sampled point subsets to construct distance 
distributions~\cite{Histogram-compare}. Wallace et 
al.~\cite{virtual-secnario-wmg, wallace2022validating} normalize these 
distance values by the maximum distance to improve robustness to 
outliers, comparing the resulting histograms via Minkowski distance. 
More recently, \cite{ali2026comprehensive} presented a benchmark study 
comparing the sensitivity of CDC, histogram matching, EMD, and ICP to 
noise, translation, and deformation, concluding that CDC is the most 
reliable of the four. 
While these geometric metrics effectively quantify point-wise positioning errors, they are fundamentally distance-based and cannot capture structural properties such as connectivity, surface topology, or spatial relationships between object components, similarly, physics-based simulation methods~\cite{hahner2022lidar, farhani20243d} rely on qualitative or downstream-task validation rather than structural metrics. Our work addresses this limitation by developing graph-based structural metrics that complement geometric validation, providing object-level diagnostics with theoretical guarantees on topological divergence.

\section{Graphs as Metrics}\label{sec:graph}

\subsection{From Point Clouds to Graphs}\label{subsec:graph-definition}
Given a point cloud $X=\{x_i\}_{i=1}^N\subset\mathbb{R}^3$, we construct an 
undirected weighted graph $G=(V,E)$ where each point is a node 
($V=\{1,\dots,N\}$) and edges encode local proximity. In practice, we form $E$ 
as the union of (i) a radius graph with threshold $\tau$ and (ii) 
$k$-nearest-neighbor connections. Each edge $(i,j)\in E$ is assigned weight
\begin{equation}
    A_{ij}=A_{ji}=
    \begin{cases}
        \|x_i-x_j\|_2, & (i,j)\in E,\\
        0,              & \text{otherwise},
    \end{cases}
    \label{equ:main_graph}
\end{equation}
yielding a weighted adjacency matrix $A\in\mathbb{R}^{N\times N}$ and degree 
matrix $D=\mathrm{diag}(d_i)$ with $d_i=\sum_j A_{ij}$. Edge weights are 
assigned as true metric distances in $\mathbb{R}^3$, rather than a 
similarity kernel such as $\exp(-\|x_i-x_j\|_2^2/\sigma^2)$, which is 
commonly used in spectral graph theory and graph neural networks to 
encode connection strength. Such kernels compress distance variation 
within a local neighborhood into near-uniform edge weights, effectively 
equalizing the contribution of nearby and distant neighbors alike. Using 
the true metric distance instead preserves a direct, monotonic penalty 
on increasing separation, so that edge weights remain geometrically 
interpretable and sensitive to spatial deviation, consistent with 
standard practice in point cloud analysis~\cite{achlioptas2018learning}.

We use the \emph{normalized graph Laplacian}
\begin{equation}
    L := I - D^{-1/2} A D^{-1/2},
    \label{eq:norm_lap}
\end{equation}
which is symmetric positive semidefinite. The dependence of spectral operators on the geometry used to construct them is not unique to graph-based analysis; related work on quantum representations of fractional Laplacians has shown that boundary assumptions can fundamentally alter the encoded operator, motivating geometry-aware spectral constructions more broadly [XX]. The Laplacian spectrum $\{\lambda_i\}_{i=0}^{N-1}$ encodes connectivity and 
structure; in particular, $\lambda_0\approx 0$ and the spectral gap 
$\mathrm{gap}(L)=\lambda_1-\lambda_0$ is sensitive to global connectivity.

\subsection{Object-Level Decomposition via Communities}\label{subsec:communities}
To obtain object-level structure, we partition each graph into communities (densely connected subgraphs)
using Louvain modularity maximization \cite{newman2006modularity,yang2016comparative}. The modularity objective is
\begin{equation}
Q=\frac{1}{2m}\sum_{i,j}\left(A_{ij}-\frac{k_i k_j}{2m}\right)\delta(c_i,c_j),
\end{equation}
where $m$ is the number of edges, $k_i$ the (weighted) degree, and $\delta(c_i,c_j)=1$ if nodes $i,j$ share
the same community. This yields community sets $\mathcal{C}_{\mathrm{real}}$ and $\mathcal{C}_{\mathrm{sim}}$
for real and simulated point clouds, respectively.

\subsection{Graph-Based Validation Metrics}\label{subsec:metrics}
We report complementary metrics at two levels: (i) \emph{scene/object presence} via community matching and
(ii) \emph{within-object structural similarity} via bounded spectral and diffusion indicators.

\subsubsection{\textbf{Community Match Rate}}\label{subsec:matchrate}
Because real and simulated scenes may contain different numbers of objects, we match communities by the spatial
proximity of their centroids (Alg.~\ref{algorithm-community}). Let $\mathcal{M}\subset\mathcal{C}_{\mathrm{sim}}$
denote the set of matched simulated communities. We report
\begin{equation}
\mathrm{Match\%}=\frac{|\mathcal{M}|}{\max(|\mathcal{C}_{\mathrm{real}}|,|\mathcal{C}_{\mathrm{sim}}|)}\times 100,
\end{equation}
where unmatched communities indicate missing or hallucinated structures.

\subsubsection{Density-Aware Chamfer Distance (CDC)}\label{subsec:cdc_here}
For each matched community pair, we compute the CDC as a geometric baseline
(as described in Sec.~\ref{sec:prior_work} is the most robust geometric baseline).
The graph metrics below provide complementary structural information not captured by geometry alone.

\subsubsection{Heat Kernel and Influence}\label{subsec:heat}
We use the heat kernel
\begin{equation}
H_t := \exp(-tL),
\label{eq:heat}
\end{equation}
which models diffusion on the graph, with diffusion time $t>0$. We define node influence by the heat-kernel
row sum,
\begin{equation}
s_i := \sum_{j} H_t[i,j],
\label{eq:influence}
\end{equation}
which emphasizes nodes that are structurally central under diffusion.

\subsubsection{\textbf{Bounded Gap Change ($r_{\mathrm{gap}}$)}}\label{subsec:rgap}
Let $\mathrm{gap}(L)=\lambda_1-\lambda_0$ denote the spectral gap. Using Weyl, the gap deviation satisfies
$|\Delta\mathrm{gap}|\le 2\|\Delta L\|_2$. We report
\begin{equation}
r_{\mathrm{gap}} := \frac{|\Delta \mathrm{gap}|}{2\|\Delta L\|_2}\le 1,
\label{eq:rgap}
\end{equation}
which is sensitive to connectivity-regime changes and near-disconnectivity.

% \subsubsection{\textbf{Bounded Diffusion Change ($r_H$)}}\label{subsec:rH}
% We quantify diffusion-level structural change by
% \begin{equation}
% r_H := \frac{\|\Delta H_t\|_2}{t\|\Delta L\|_2}\le 1,
% \qquad \Delta H_t := H_t(L_B)-H_t(L_A),
% \label{eq:rH}
% \end{equation}
% which normalizes the heat-kernel deviation by the perturbation scale.

\subsubsection{\textbf{Bounded Spectral Change ($r_\lambda$)}}
Let $L_A$ and $L_B$ denote the reduced normalized Laplacians of a
matched real--simulated community pair after heat-kernel node selection
and Hungarian spatial alignment. Define
\begin{equation}
\Delta L := L_B - L_A .
\end{equation}
Let $\{\lambda_i(L_A)\}_{i=0}^{n_{\mathrm{keep}}-1}$ and
$\{\lambda_i(L_B)\}_{i=0}^{n_{\mathrm{keep}}-1}$ denote the ordered
eigenvalues of $L_A$ and $L_B$, respectively. We define the bounded
spectral-change metric as
\begin{equation}
r_\lambda :=
\frac{
\max_i |\lambda_i(L_B)-\lambda_i(L_A)|
}{
\|\Delta L\|_2
}
\le 1 .
\end{equation}
The bound follows from Weyl's inequality for symmetric matrices,
\begin{equation}
\max_i |\lambda_i(L_B)-\lambda_i(L_A)|
\le
\|L_B-L_A\|_2 .
\end{equation}
Thus, $r_\lambda$ measures the largest normalized eigenvalue shift
across the full Laplacian spectrum and provides a bounded indicator of
within-object structural change.

\subsection{Practical Pipeline}\label{subsec:pipeline}
An overview of the practical pipeline is shown in the top panel of
Fig.~\ref{fig:intro}; its steps are detailed below.
\subsubsection{Community Matching}\label{subsec:community_matching}
Algorithm~\ref{algorithm-community} matches communities between real and simulated graphs based on centroid
proximity, enabling object-level comparison when the number of detected communities differs.

\begin{algorithm}[h!]
\caption{Community Matching Algorithm}
\begin{algorithmic}[1]
\Require Graphs $G_1$ (real), $G_2$ (simulated), threshold $\delta$
\Ensure Percentage of matched communities

\State \textbf{Detect Communities:} Apply Louvain algorithm to $G_1$ and $G_2$ (fixed seed), yielding sets
$\mathcal{C}_1$ and $\mathcal{C}_2$

\State \textbf{Compute Community Centroids:}
\For{each community $c \in \mathcal{C}_1 \cup \mathcal{C}_2$}
    \State $\mathrm{Center}(c) = \frac{1}{|c|}\sum_{v\in c} x_v$
\EndFor

\State \textbf{Greedy Matching:} $\mathrm{Matched}\leftarrow 0$, $\mathcal{M}\leftarrow\emptyset$
\For{each centroid $p_1$ from $\mathcal{C}_1$}
    \State $p_2^\star = \arg\min_{p_2\in \mathcal{C}_2\setminus\mathcal{M}} \|p_1-p_2\|_2$
    \If{$\|p_1-p_2^\star\|_2<\delta$}
        \State $\mathcal{M}\leftarrow \mathcal{M}\cup\{p_2^\star\}$, $\mathrm{Matched}\leftarrow \mathrm{Matched}+1$
    \EndIf
\EndFor

\State \textbf{Match Percentage:}
$\mathrm{Match\%}=\frac{\mathrm{Matched}}{\max(|\mathcal{C}_1|,|\mathcal{C}_2|)}\times 100$

\State \Return $\mathrm{Match\%}$
\end{algorithmic}
\label{algorithm-community}
\end{algorithm}

\subsubsection{Heat-Kernel Node Selection for Density Mismatch}\label{subsec:node_selection}
Matched communities often have different point counts ($n_{\mathrm{real}}\neq n_{\mathrm{sim}}$), making
direct spectral comparison ill-posed. We therefore reduce each matched community to a common size
$n_{\mathrm{keep}}=\min(n_{\mathrm{real}},n_{\mathrm{sim}})$ using heat-kernel influence scores (Alg.~\ref{algo2}),
then compute the structural metrics in Sec.~\ref{subsec:metrics} on the reduced subgraphs.

\begin{algorithm}[h!]
\caption{Heat Kernel Node Selection for Density-Mismatched Communities}
\begin{algorithmic}[1]
\Require Matched communities from $G_{\mathrm{real}}$ and $G_{\mathrm{sim}}$, diffusion time $t$
\Ensure Reduced subgraphs with equal node counts and their Laplacians

\State \textbf{Extract Community Laplacians:} Obtain $L_{\mathrm{real}}$ and $L_{\mathrm{sim}}$ (Eq.~\ref{eq:norm_lap})

\State \textbf{Compute Heat Kernels:}
$H_{\mathrm{real}}=\exp(-tL_{\mathrm{real}})$,\;\;
$H_{\mathrm{sim}}=\exp(-tL_{\mathrm{sim}})$

\State \textbf{Compute Influence Scores:} $s_i=\sum_j H[i,j]$ for each graph (Eq.~\ref{eq:influence})

\State \textbf{Select Top Nodes:} Set $n_{\mathrm{keep}}=\min(n_{\mathrm{real}},n_{\mathrm{sim}})$ and retain the top
$n_{\mathrm{keep}}$ nodes by influence in each community

\State \textbf{Build Reduced Subgraphs:} Construct $G'_{\mathrm{real}}$ and $G'_{\mathrm{sim}}$ on the selected nodes,
preserving induced edges

\State \Return Reduced subgraphs $G'_{\mathrm{real}}$, $G'_{\mathrm{sim}}$ and their Laplacians
\end{algorithmic}
\label{algo2}
\end{algorithm}

After node selection, we compute the reduced normalized Laplacians and report $r_\lambda$, $r_{\mathrm{gap}}$,
as per-community structural similarity indicators,
alongside CDC as a geometric baseline.

\subsection{Data Collection and Graph Build-up}

The ground-truth point clouds were collected with a vehicle-mounted Velodyne Puck (VLP--32C) LiDAR sensor. The procedure used to construct the VTE from the LiDAR data is reported in the same references; briefly, we acquired synchronized LiDAR, IMU, and camera measurements and estimated vehicle odometry using a LiDAR--inertial SLAM pipeline following the simulation-based approach of~\cite{ali2026comprehensive, ali2025virtual}. The resulting registered point clouds were then used to generate a 3D mesh of the environment, which was imported into CARLA~\cite{carla} to create a one-to-one VTE. This setup enables direct comparison between simulated and real LiDAR scans captured in the same scene geometry.

For each real and simulated scan, we construct a graph using Eq.~\ref{equ:main_graph} by connecting each point to its $k$ nearest neighbors and additionally enforcing a radius threshold. These parameters were selected empirically to reflect the local sampling density of meaningful object surfaces, where each point typically has on the order of tens of neighbors within close spatial proximity. Since road data typically forms a ring-shaped structure with a well-defined geometry \cite{ali2026comprehensive}, we apply a Z-value cut-off to exclude the road surface and focus on objects above ground for graph comparison.

The Z-threshold is determined using Principal Component Analysis (PCA). Initially, an \textit{a-priori} threshold is set, and the variance loss due to this cut-off is compared to the case where no Z-threshold is applied. The threshold is then iteratively adjusted to ensure that the variance loss remains within 1\% of the total variance.

Ground removal is a widely used technique \cite{zeng2024can}, and similar PCA-based approaches for defining road surfaces have been introduced, as demonstrated by \cite{bartels2010threshold}.

We then perform community detection on both the real and simulated graphs using Algorithm~1.

\section{Results and Analysis}

\subsection{Controlled Perturbation Analysis}
\label{sec:perturbation}

We evaluate the sensitivity and robustness of the proposed metrics under
three perturbation types, Gaussian noise, point dropout, and rigid
transforms, applied to four object classes: vegetation, car, tree, and
wall. These classes were selected to span a range of geometric
complexity, from irregular branching structures (vegetation, tree) to
curved compact surfaces (car) and large planar structures (wall).
Fig.~\ref{fig:petrubation-plots}(a) shows the selected communities (based on following Algorithm~1, to generate the communities, and Fig.~\ref{fig:petrubation-plots}(b) and Fig.~\ref{fig:petrubation-plots}(c)
indicate resulting metric curves.

\subsubsection{Rigid Transforms}
Applying translations ($t_{\mathrm{mag}} = 0.05$\,m) and rotations
($\theta = 2^\circ$) yields $r_\lambda < 0.01$ and $r_{\mathrm{gap}} < 0.0001$ across all four object classes, confirming that the metrics are
invariant to global pose changes and respond only to structural
deformation. This is the expected behaviour: a rigid transform preserves
point proximity relationships and therefore leaves the graph Laplacian
unchanged. 

\subsubsection{Gaussian Noise}
We add i.i.d.\ Gaussian noise with $\epsilon \in \{10^{-3}, 10^{-2},
10^{-1}\}$ and repeat each condition twenty times; shaded bands in
Fig.~\ref{fig:petrubation-plots}(b,c) show $\pm 1$ std.

For \textbf{vegetation}, both $r_\lambda$ and $r_{\mathrm{gap}}$ grow
slowly with $\epsilon$, indicating that the irregular branching
structure absorbs small displacements without significant spectral
drift. The \textbf{tree} community shows a near-flat response across
all noise levels: its large spatial extent produces a low edge-rewiring
rate per unit noise, resulting in minimal Laplacian change. The
\textbf{wall} exhibits a moderate, steady increase in $r_\lambda$,
consistent with the sensitivity of uniform planar connectivity to
perturbations that disrupt edge regularity. The \textbf{car} shows a
sharp jump in $r_\lambda$ between $\epsilon = 10^{-2}$ and $10^{-1}$
(Fig.~\ref{fig:petrubation-plots}b): at this scale, noise rewires edges
on the curved compact surface, causing a sudden spectral shift confirmed
by the corresponding rise in $r_{\mathrm{gap}}$
(Fig.~\ref{fig:petrubation-plots}c). Across all classes, the metrics
remain small for $\epsilon \leq 10^{-2}$, demonstrating robustness to
typical sensor measurement noise while remaining sensitive to
structurally significant deformation.

\subsubsection{Point Dropout}
\begin{figure}
    \centering
    \includegraphics[width=1\linewidth]{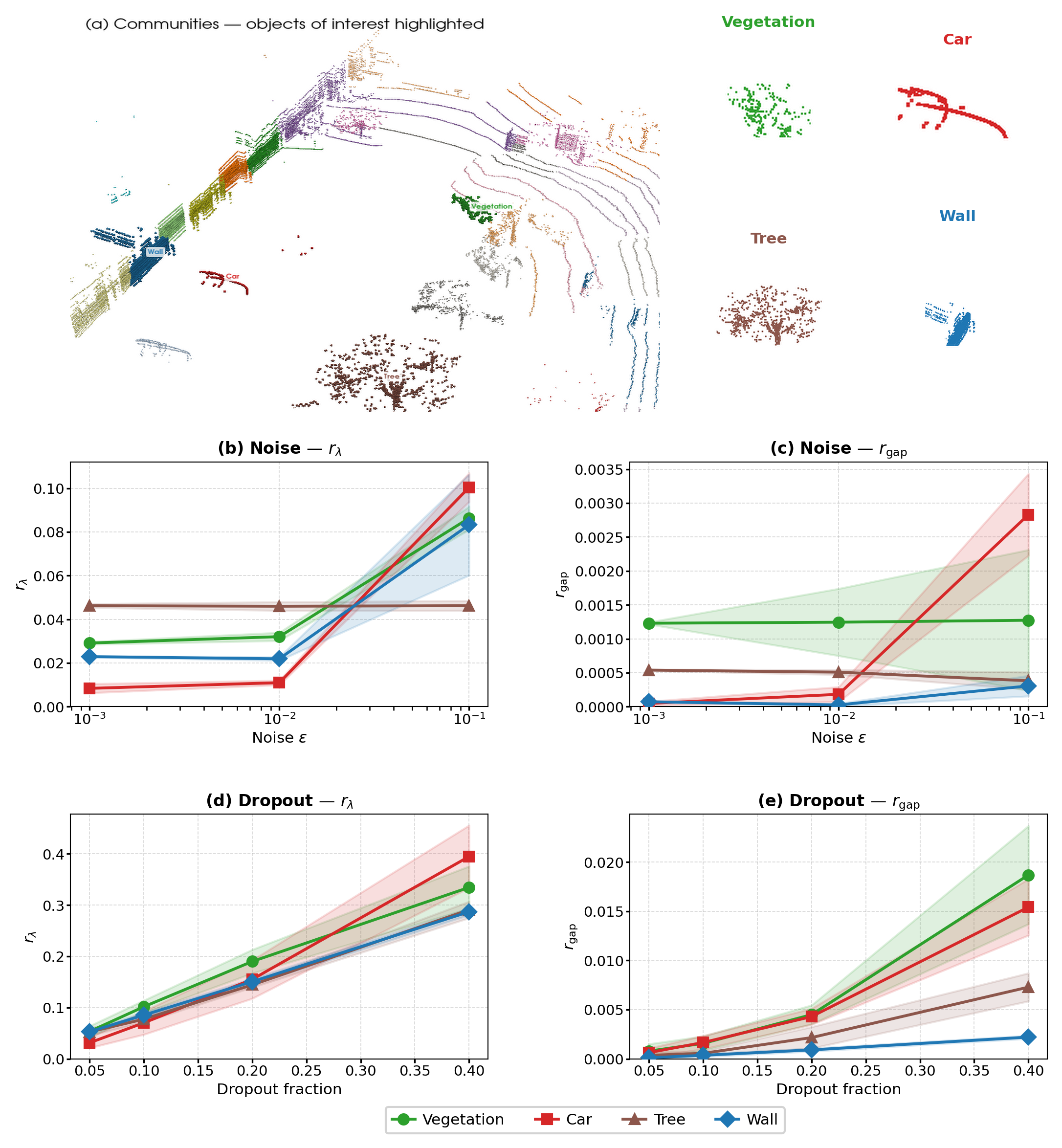}
    \caption{Controlled perturbation analysis. (a)~Scene communities
    (left) and the four selected object subgraphs (right): vegetation
    (green), car (red), tree (brown), wall (blue). (b,c)~Effect of
    Gaussian noise on $r_\lambda$ and $r_{\mathrm{gap}}$; shaded bands
    show $\pm 1$ std over five seeds. (d,e)~Effect of point dropout.
    Car and vegetation are most sensitive; wall is most robust due to
    connectivity redundancy in planar structures. }
    \label{fig:petrubation-plots}
\end{figure}

Dropout produces a substantially stronger effect than noise across all
object classes (Fig.~\ref{fig:petrubation-plots}d,e). Both $r_\lambda$ and $r_{\mathrm{gap}}$ increase monotonically with dropout fraction:
car and vegetation are most sensitive, reaching $r_\lambda \approx
0.40$ and $0.35$ respectively at 40\% dropout, while wall remains the
most robust with $r_{\mathrm{gap}} < 0.004$ throughout. This reflects
connectivity redundancy in large planar structures, removing points from
a wall leaves many alternative paths intact, whereas removing points
from a curved or branching object readily disconnects critical edges.
These object-level differences confirm that the metrics capture
geometry-dependent structural sensitivity rather than a uniform global
response.

Point dropout also serves as a controlled proxy for the density mismatch
inherent in real-versus-simulated comparisons. A simulator may
undersample surfaces relative to the real sensor (analogous to dropout),
or generate spurious points absent in the real scan (hallucination). In
our real-versus-simulated experiments, the
density ratio between simulated and real communities varies
substantially across objects: $n_{\mathrm{sim}}$ ranges from less than
one quarter to more than four times $n_{\mathrm{real}}$ depending on
the object surface and viewing angle. This motivates the heat kernel
node selection, which reduces each matched community to
$n_{\mathrm{keep}} = \min(n_{\mathrm{real}}, n_{\mathrm{sim}})$ before
computing spectral metrics, ensuring that structural comparison is never
confounded by density mismatch regardless of its direction or magnitude.

The slower growth of $r_{\mathrm{gap}}$ compared with $r_\lambda$ can be
understood from the fact that the spectral gap is a low-dimensional summary of
\textbf{global connectivity}. Since the analysis is performed only after object-level
community matching, the compared subgraphs typically already share the same
coarse connectivity regime. Therefore, perturbations that modify \textbf{local edge
structure} or surface connectivity may shift the broader Laplacian spectrum,
and hence increase $r_\lambda$, without substantially changing the first
nonzero eigenvalue. In this sense, $r_{\mathrm{gap}}$ is mainly sensitive to
connectivity-regime changes such as bottlenecks or near-disconnections, whereas
$r_\lambda$ captures more distributed internal structural deformation.

\subsubsection{\textbf{Global Point Cloud Perturbation}}

For the global perturbation analysis, we examine how the percentage of matched communities between the real and perturbed point clouds changes after community detection. The results are summarized in Table~\ref{tab:community_perturbation_summary_reduced}.

For additive Gaussian noise, the percentage of matched communities remains high across all tested noise levels. Although a slight degradation is observed as the noise level increases, the match rate remains above 95\% even for the largest perturbation considered. At the same time, the centroid distance $d_{\mathrm{cent}}$ increases with noise magnitude, indicating that the spatial locations of matched communities become more displaced. The eigenspectrum distance $d_{\lambda}$ also increases gradually, suggesting a mild perturbation of the internal graph structure, although the increase remains relatively small overall. These results indicate that the community structure is largely preserved under Gaussian noise.

A similar trend is observed for rigid transformations. For the tested small translations and rotations, the matched-community rate remains equal to 100\%, showing that these perturbations do not disrupt the community correspondence. The eigenspectrum distance $d_{\lambda}$ remains essentially negligible, confirming that the structural graph properties are preserved. As expected, however, $d_{\mathrm{cent}}$ increases under rigid transformations because the community centroids are physically displaced by translation and rotation, even though the underlying graph structure remains unchanged.

In contrast, point dropout has a much stronger effect. As the dropout rate increases, the matched-community rate decreases from approximately 91.5\% at $p=0.05$ to 85.4\% at $p=0.40$. This reduction is accompanied by a substantial increase in both $d_{\mathrm{cent}}$ and $d_{\lambda}$, indicating not only larger spatial displacement between matched communities but also greater structural mismatch in their spectral signatures. This observation is consistent with the earlier local analysis, where dropout was found to be the most disruptive perturbation. Overall, these results suggest that the proposed graph-based representation is robust to Gaussian noise and small rigid transformations, while remaining sensitive to structural degradation caused by point removal.

\begin{table*}[t]
\centering
\footnotesize
\setlength{\tabcolsep}{5pt}
\renewcommand{\arraystretch}{1.08}
\begin{tabular}{l|c c|c c c}
\hline
\textbf{Perturb.} & \textbf{Param 1} & \textbf{Param 2} & Match rate & $d_{\mathrm{cent}}$ & $d_{\lambda}$ \\
\hline
\multicolumn{6}{c}{\textbf{Noise sweep, mean $\pm$ std}}\\
\hline
Noise & $\epsilon=1.0\times10^{-3}$ & -- & $0.9593\pm0.0130$ & $0.0202\pm0.0323$ & $0.0087\pm0.0062$ \\
Noise & $\epsilon=1.0\times10^{-2}$ & -- & $0.9533\pm0.0400$ & $0.0334\pm0.0271$ & $0.0209\pm0.0038$ \\
Noise & $\epsilon=1.0\times10^{-1}$ & -- & $0.9529\pm0.0338$ & $0.0733\pm0.0586$ & $0.0435\pm0.0112$ \\
\hline
\multicolumn{6}{c}{\textbf{Dropout sweep, mean $\pm$ std}}\\
\hline
Dropout & $p=0.05$ & -- & $0.9150\pm0.0528$ & $0.2349\pm0.0765$ & $0.1003\pm0.0176$ \\
Dropout & $p=0.20$ & -- & $0.8837\pm0.0730$ & $0.3753\pm0.0945$ & $0.3001\pm0.0228$ \\
Dropout & $p=0.40$ & -- & $0.8543\pm0.0233$ & $0.6269\pm0.1101$ & $0.7855\pm0.0572$ \\
\hline
\multicolumn{6}{c}{\textbf{Rigid transform sweep, mean $\pm$ std}}\\
\hline
Rigid & $t_{\mathrm{mag}}=0.00$ & $\theta=0.5^\circ$ & $1.0000\pm0.0000$ & $0.2063\pm0.0000$ & $7.213\times10^{-6}\pm0$ \\
Rigid & $t_{\mathrm{mag}}=0.00$ & $\theta=1.0^\circ$ & $1.0000\pm0.0000$ & $0.4126\pm0.0000$ & $1.576\times10^{-7}\pm0$ \\
Rigid & $t_{\mathrm{mag}}=0.05$ & $\theta=2.0^\circ$ & $1.0000\pm0.0000$ & $0.8299\pm0.0049$ & $4.494\times10^{-6}\pm5.511\times10^{-6}$ \\
\hline
\end{tabular}
\caption{Community-level perturbation stability summary (mean $\pm$ std over seeds) for noise, dropout, and rigid transforms. Reported metrics are matched-community rate, mean centroid distance $d_{\mathrm{cent}}$, and mean eigenspectrum distance $d_{\lambda}$.}
\label{tab:community_perturbation_summary_reduced}
\end{table*}

\subsection{Real experiments}\label{sec:real experiments}

Having established the robustness of the proposed graph-based metrics under controlled scan perturbations, we next evaluate their behavior on real and simulated LiDAR scans generated in CARLA. Across 50 scan pairs, the algorithm produced 1,067 matched communities, with a one-to-one community match rate of $59.4\% \pm 0.066$ (Algorithm~1). 

\begin{figure*}
    \centering
    \includegraphics[width=0.7\linewidth]{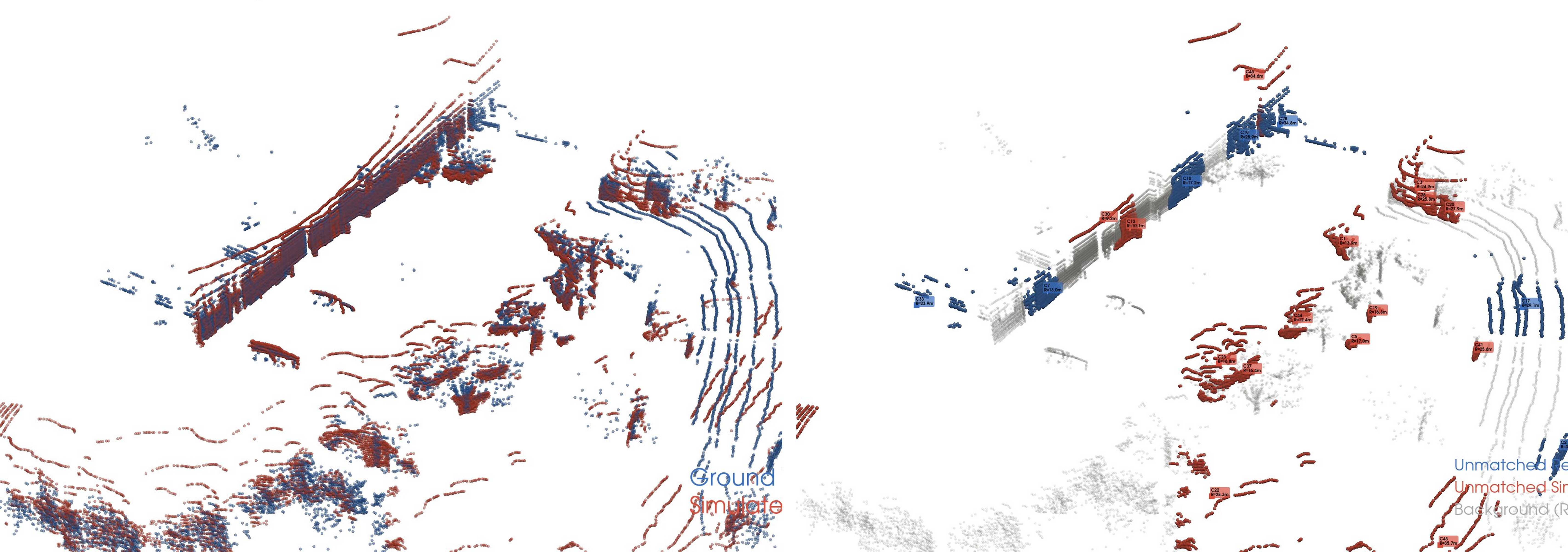}
    \includegraphics[width=0.4\linewidth]{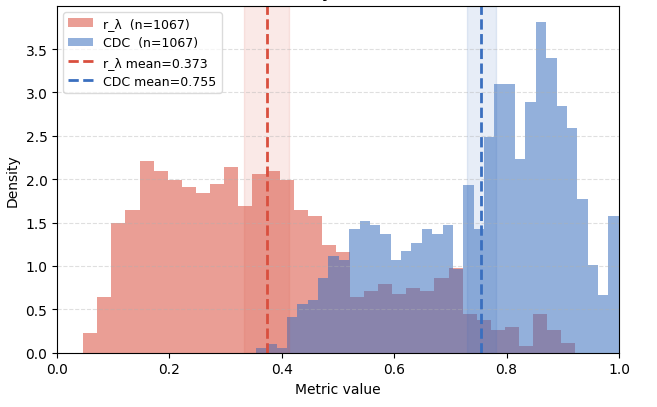}
    \includegraphics[width=0.4\linewidth]{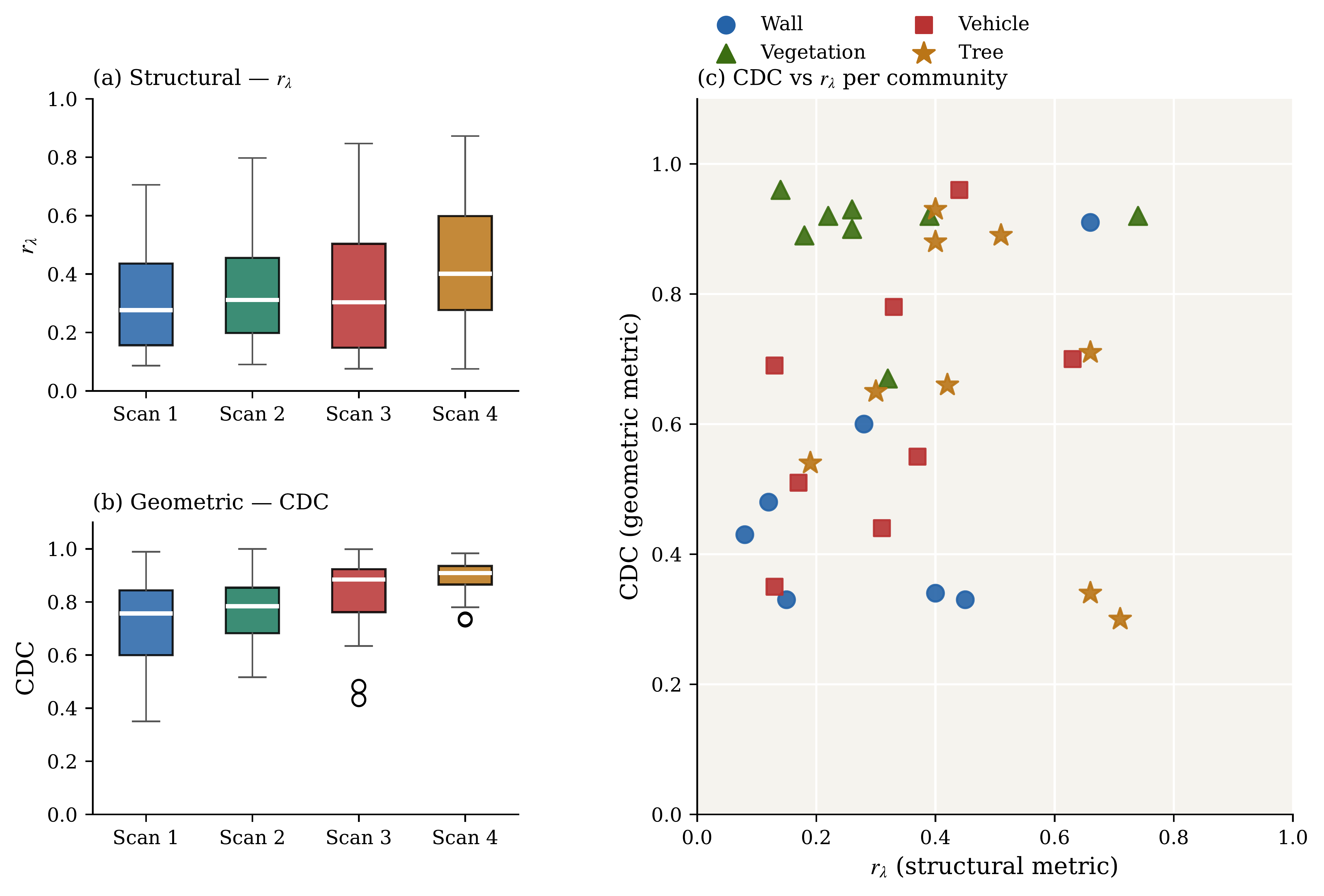}
    \caption{Left panel: The selected communities are color coded according to
    $r_\lambda$. Right panel: Four representative objects are selected to
    compare graph-structural similarity ($r_\lambda$) with geometric similarity
    (CDC).}
    \label{fig:real_experiments_stat}
\end{figure*}

To better understand the source of these mismatches and determine whether they stem from limitations of the community matching algorithm or reflect genuine geometric discrepancies, we examine the representative example shown in Top Panel of Fig.~\ref{fig:real_experiments_stat}. On the left, overlays the real (shown in blue) and simulated point clouds (shown in red). As is clearly visible, many mismatches arise because real (ground-truth) point clouds exist in regions where no corresponding simulated points are found (see Right Panel where the no-matches are color coded). The median distance of unmatched communities from the sensor origin was $29$~m, suggesting that mismatches occur predominantly in more distant parts of the scene, where mesh construction and ray tracing introduce greater inaccuracies. Consequently, for many locations, the simulation simply produces \textbf{no points} to match against. 

A second category of mismatch involves hallucinated structures in the simulated scan: regions where CARLA's ray tracing generates an excess of points that have no real-world counterpart, while the actual point cloud is far sparser and more dispersed. As shown in the right panel, unmatched simulated communities frequently correspond to such hallucinated structures, which tend to absorb nearby real points into larger simulated communities rather than forming clean one-to-one correspondences. Across multiple scans, this behavior was especially pronounced for trees with dense foliage and, more broadly, for distant objects. Conversely, unmatched communities in the real scans typically correspond to structures that are absent from the simulation entirely. 

These observations are consistent with the controlled perturbation experiments, where point-level mismatches, including dropout and hallucination, were shown to disrupt community correspondence. In the real experiments, however, this effect is more pronounced: distant foliage and tree structures are frequently missing from the simulated scans, while several unmatched simulated communities appear to arise from hallucinated geometry, likely introduced by limitations in the meshing or simulation process

Quantitatively, across the 50 scans, CDC averaged $0.75 \pm 0.02$, while 
$r_\lambda$ averaged $0.37 \pm 0.03$ and $r_{\mathrm{gap}}$ averaged 
$0.028 \pm 0.014$. Consistent with the controlled perturbation study, 
the small $r_{\mathrm{gap}}$ indicates that matched communities 
generally preserve the same \textbf{coarse connectivity regime}, with no 
major bottleneck formation or near-disconnection. In contrast, the 
larger and more variable $r_\lambda$ values reveal \textbf{internal 
structural changes} within the matched objects. The bottom-left panel 
of Fig.~\ref{fig:real_experiments_stat} shows the distribution of 
$r_\lambda$ and CDC across the 1,067 matched objects from all 50 scans: 
$r_\lambda$ spans a wider range with a lower, broader peak, whereas CDC 
is concentrated in a narrower range with a sharper, higher 
peak, indicating that CDC is less able to discriminate 
differences between matched communities.

To further examine this behavior, we selected four scans separated by
more than one minute in acquisition time to ensure scene diversity. The bottom-middle panel in Fig.~\ref{fig:real_experiments_stat}
panel shows box plots of CDC and $r_\lambda$ across all
objects in these four scans. Consistent with the aggregate analysis, CDC
remains high and only weakly dispersed, whereas $r_\lambda$ shows
substantially larger variability. The bottom-right panel further compares
CDC and $r_\lambda$ for selected object classes, namely trees, walls,
vehicles, and vegetation, across the four scans, corresponding to a total
of 56 objects.

\begin{figure*}
    \centering
 \includegraphics[width=0.40\linewidth]{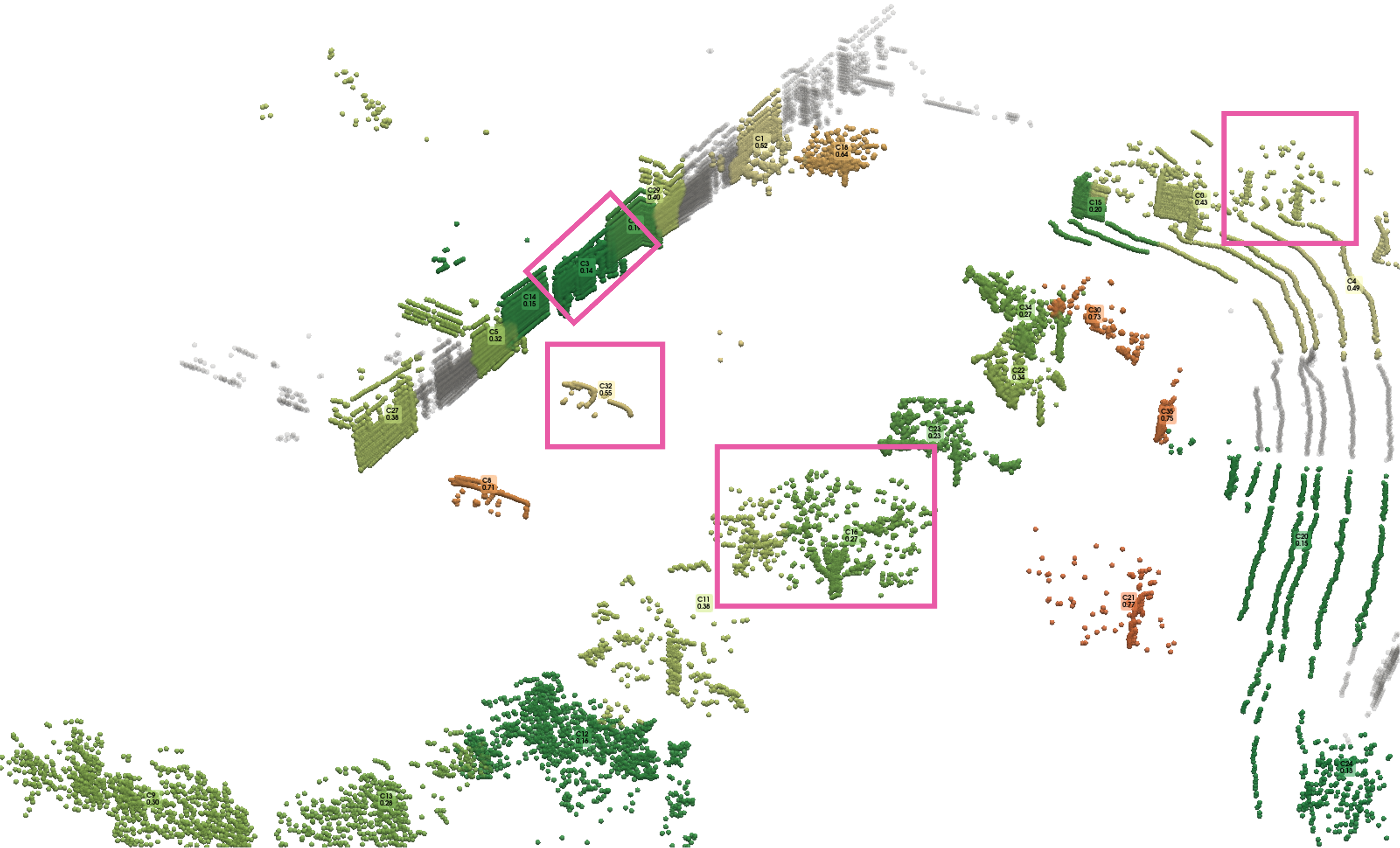}
    \includegraphics[width=0.30\linewidth]{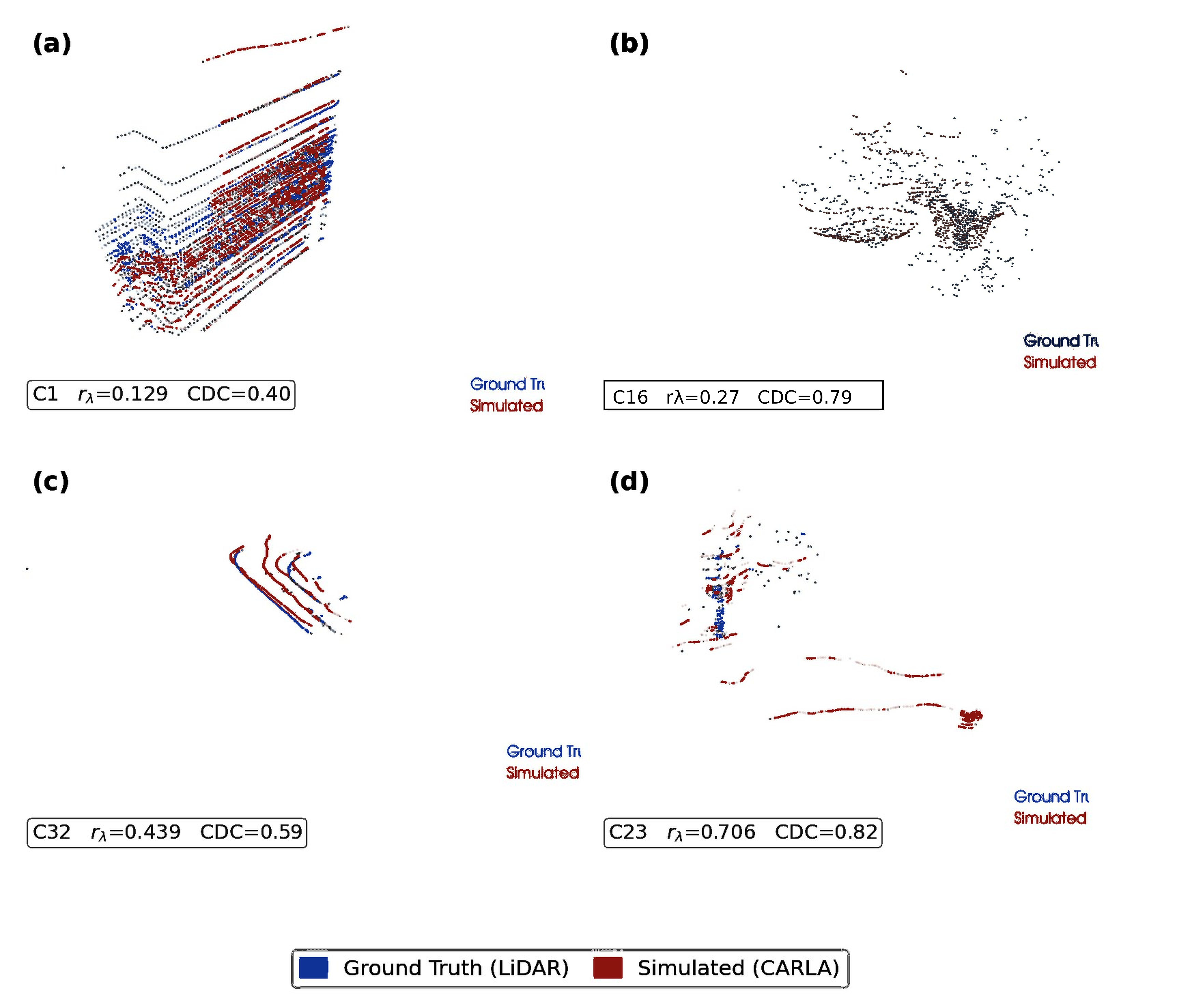}
\caption{Left panel: The selected communities are color coded according to
$r_\lambda$. Right panel: Four representative objects are selected to compare
graph-structural similarity ($r_\lambda$) with geometric similarity (CDC).}
    \label{fig:real_experiment_scan_1}
\end{figure*}

A representative example, based on one scan, is shown in the left panel 
of Fig.~\ref{fig:real_experiment_scan_1}, where communities are 
color-coded according to their $r_\lambda$ value. In subplot (a), 
corresponding to a wall, CDC $=0.40$ and $r_\lambda=0.13$ both indicate 
small discrepancy, showing close agreement between the two metrics. 
Note that CDC does not take values below 0.4 in our data, so this value 
represents its lower end, i.e., the best achievable agreement, whereas 
$r_\lambda$ spans a wider range and its value of 0.13 likewise indicates 
a good match.

Subplots (b) and (d) correspond to two trees at different distances 
from the sensor. In (b), the tree closer to the LiDAR exhibits a 
relatively small $r_\lambda=0.27$ but a large CDC value of $0.79$, 
indicating that its graph structure is largely preserved even though 
the corresponding CDC remains high, as also reflected in the right 
panel. In (d), the more distant tree shows both a larger CDC of $0.82$ 
and a larger $r_\lambda$ of $0.70$. Together, these two subplots suggest 
that $r_\lambda$ better distinguishes cases in which structural 
similarity is preserved despite geometric differences: many trees near 
the sensor retain similar graph structure, while CDC assigns uniformly 
high dissimilarity with little discrimination. This is consistent with 
the box plots, which show that the variance of $r_\lambda$ across 
objects within a scan is much wider than that of CDC.

Subplot (c) shows a vehicle, one of the most safety-critical object 
classes. Here, CDC $=0.56$ and $r_\lambda=0.43$ agree reasonably well, 
with both the structural and geometric metrics indicating moderate 
differences.

This trend was observed consistently across the four selected scans and 
more broadly across the full set of 50 scans. It helps explain why the 
mean CDC over all scans is substantially larger than the mean 
$r_\lambda$: although the underlying graph structure often remains 
similar, CDC still indicates large geometric discrepancies.

% \begin{figure}[h]
%     \centering
%     \includegraphics[width=1\linewidth]{fig/overlapped_unmatched_sidebyside.png}
%     \caption{Left panel: the overlay of real (ground truth) and simulated point clouds. Real point clouds are shown in blue, while simulated is shown in red. Right Panel: the unmatched communities are shown, the unmatchede from simuklated are shown in red, and unmatched from real in blue.  }
%     \label{fig:unmatch}
% \end{figure}

%and the simulated points clouds for this work are taken from [REF Syed'd work] where the process has explained in detail. 
\subsubsection{Pipeline Timing}
Although graph construction and community detection introduce non-trivial computational overhead, this is acceptable given that fidelity evaluation is performed offline. All experiments were run on Google Colab (high-RAM CPU runtime), requiring no specialized hardware. The four primary pipeline stages and their per-scan runtimes are: community detection, 544~ms; Laplacian computation across all subgraphs, 596~ms; heat kernel computation, 1{,}505~ms; and Hungarian spatial alignment across all subgraphs, 1{,}532~ms. Including all intermediate steps, the total end-to-end runtime per scan pair is approximately 7{,}120~ms.

\subsection{Ablation Study}
The graphs are constructed by connecting each point to its $k$ nearest neighbors, subject to an additional radius threshold $\tau$. Both parameters are treated as hyperparameters to be tuned, and their primary influence is on the community matching process. Smaller values of $k$ yield sparser graphs, whereas larger values produce denser, more connected graphs. In the sparser regime, community matching becomes more sensitive, as small spatial displacements can lead to substantially different graph structures. For the heat kernel, the diffusion time $t$ is also a hyperparameter; however, as expected, it has negligible influence on either community matching or $r_{\lambda}$. Similarly, the community matching threshold $\delta$ affects the matching process itself but does not influence $r_{\lambda}$.

\begin{figure}[h]
    \centering
    \includegraphics[width=0.8\linewidth]{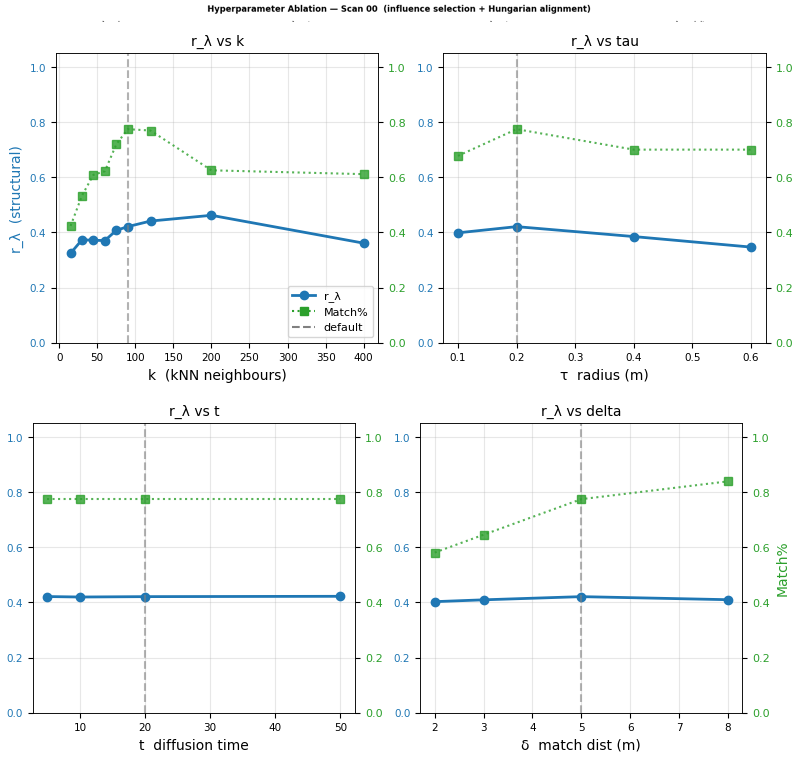}
    \caption{Changes in $r_\lambda$ (blue) and community match percentage
    (green) for different hyperparameter values.}
    \label{fig:ablation}
\end{figure}

To select appropriate values, we evaluated $r_{\lambda}$ and the community matching percentage across a range of parameter combinations, arriving at optimal values of $k = 90$, $\tau = 0.2$, and $\delta = 5$. Consistent with theoretical expectations, the choice of diffusion time $t$ had no meaningful effect on either community detection or $r_{\lambda}$ (see Fig.~\ref{fig:ablation}).

\section{Conclusion}

This work introduced a graph-based structural evaluation framework to complement geometric metrics in the comparison of real and simulated LiDAR point clouds. The central premise is that geometry and structure capture distinct aspects of simulation fidelity: a point cloud may approximate the spatial layout of a real scan while differing substantially in connectivity, or preserve structural organization despite geometric displacement. In this framework, CDC captures geometric similarity and $r_\lambda$ captures structural similarity.

Controlled perturbation experiments confirmed that $r_\lambda$ is invariant to rigid transforms and sensor noise, while remaining sensitive to structural deformation from point dropout and hallucination, the same artifacts most prevalent in real simulator outputs. In real-data experiments, the low community match rate was largely attributable to systematic simulator deficiencies: hallucinated structures and missing returns from distant regions, rather than failures of the matching procedure itself.

The key finding is that CDC and $r_\lambda$ are complementary rather than redundant. CDC remained consistently high and offered limited discrimination across fidelity levels, whereas $r_\lambda$ exhibited a substantially larger dynamic range. For trees and vegetation, the two metrics frequently decoupled: successfully matched communities often retained similar graph structure even when CDC indicated large geometric discrepancy. For compact rigid objects such as vehicles, the two metrics were generally consistent.

The framework is model-free and requires no labeled data, making it well-suited as a trustworthy validation metric. We propose joint reporting of $r_\lambda$ and CDC as a richer diagnostic for VTE validation, revealing failure modes that neither metric captures in isolation.

\bibliographystyle{IEEEtran}
\begingroup
\let\footnotesize\scriptsize
\bibliography{IEEEfull}
\endgroup

\end{document}